# Gradient Guided Hypotheses: A unified solution to enable machine learning models on scarce and noisy data regimes

Paulo Neves[1,2], Jörg K. Wegner[1], Philippe Schwaller[2,3]

1 In-Silico Discovery (ISD), Janssen Research & Development, Janssen Pharmaceutica N.V
2 ISIC, EPFL, Laboratory of Artificial Chemical Intelligence (LIAC), Lausanne, 1050, VD, Switzerland
3 National Centre of Competence in Research (NCCR) Catalysis, Ecole Polytechnique Fédérale de Lausanne (EPFL), Lausanne, Switzerland

## ABSTRACT

Ensuring high-quality data is paramount for maximizing the performance of machine learning models and business intelligence systems. However, challenges in data quality, including noise in data capture, missing records, limited data production, and confounding variables, significantly constrain the potential performance of these systems. In this study, we propose an architecture-agnostic algorithm, Gradient Guided Hypotheses (GGH), designed to address these challenges. GGH analyses gradients from hypotheses as a proxy of distinct and possibly contradictory patterns in the data. This framework entails an additional step in machine learning training, where gradients can be included or excluded from backpropagation. In this manner, missing and noisy data are addressed through a unified solution that perceives both challenges as facets of the same overarching issue: the propagation of erroneous information.

Experimental validation of GGH is conducted using real-world open-source datasets, where records with missing rates of up to 98.5% are simulated. Comparative analysis with state-of-the-art imputation methods demonstrates a substantial improvement in model performance achieved by GGH. Specifically in very high scarcity regimes, GGH was found to be the only viable solution. Additionally, GGH's noise detection capabilities are showcased by introducing simulated noise into the datasets and observing enhanced model performance after filtering out the noisy data. This study presents GGH as a promising solution for improving data quality and model performance in various applications.

## 1. Introduction

Capturing and representing the real world through numerical vector representations presents multiple complex challenges. Two of the most critical are missing data and the inaccuracy of the data collected, a phenomenon well-documented in the literature [1] [2] [3] [4].

Inaccurate or noisy data is a widely recognized problem in data-driven methodologies, often leading to a degradation in model performance [5]. Even high performance on standard benchmarks, such as the ImageNet validation set, may not accurately reflect a model's effectiveness in real-world scenarios, as highlighted by Northcutt et al. [6].

The challenge of missing data can be subdivided into two primary forms: a variable can be partially missing, or completely omitted from a dataset. If the latter has a causal link to the dependent variable, it is denominated as a confounding variable. Identifying and addressing the impact of such variables is crucial across various domains, from health research [7] [8] to economics [9]. If data is accurately captured without missing values, the residual "noise" can be attributed to the influence of confounding variables. Therefore, distinguishing between data instances whose outcome is fully explained by the recorded variables and those that are not, may offer a unified solution to the challenges of missing and inaccurate data.

The objective of this publication is twofold: to introduce a novel technique to tackle the dual issues of missing and inaccurate data, and to inspire a new paradigm. In this paradigm, low-cost data production efforts could target confounding variables to scarcely populate the training data, which combined with GGH would enable models to achieve some understanding of the confounding dynamics.

## 2. State of the art
**2.1 Noise Detection**

Addressing noise in datasets remains a critical challenge in the fields of machine learning (ML) and data analysis, as it can significantly affect the accuracy and efficiency of the models and insights that can be extracted from the data [10]. Noise can be subdivided into two categories, feature noise and label noise, which most commonly occur when features or label instances are incorrectly captured or assigned.

The first noise detection methods can be traced back to the statistical techniques pioneered by Karl Pearson in the early 20th century. Notable among these is the Z-score analysis, which assesses data point anomalies by its distance to the mean in standard deviations. Another method from the same period, the Interquartile Range, employs a similar approach but uses a percentile-based threshold instead of standard deviations.

Over time subsequent research has been directed towards generalist label noise detection methods due to their higher impact. However, domain-specific methods and generalist feature noise detection have also made noteworthy strides. Early examples of such domain-specific metrics are the signal-to-noise ratio introduced with the development of electrical communication systems. More recently, other domain-specific methods were introduced, such as those applied to genomic quality scores in bioinformatics. Noteworthy strides have also been made in feature noise detection, such as the pairwise attribute noise detection algorithm (PANDA) which surpassed nearest-neighbor based techniques [11] in 2007 before the popularization of ML methods. As more computing power became available, ensemble-based methods emerged, leveraging existing techniques to form more robust solutions. These ensemble approaches have consistently demonstrated superior efficacy in identifying noisy instances, thus providing more accurate and appropriate methodologies with uncertainty estimation for handling noisy datasets[12].

In parallel to the development of noise detection techniques, there has also been progress with inherently noise-resilient algorithms such as Random Forest (RF) and Neural-Networks (NN). These algorithms have shown to some extent the ability to address random systematic label noise, offering an indirect partial solution to the noise challenge. This approach inadvertently constrains the choice of algorithms, excluding alternatives like Support Vector Machines (SVM) and logistic regression which are sensitive to noise[13]. It should be noted, of course, that at higher noise thresholds, even the performance of RF and NN are severely impacted. This vulnerability is especially evident in classification tasks where noise presence is binary. In such scenarios, an incorrect class label translates to a complete inaccuracy, as opposed to a percentage deviation that might occur with continuous variables.

More recently, ML techniques have become popular tools for this task, we highlight the following which were utilized in combination with our method:
- Density-Based Spatial Clustering of Applications with Noise DBSCAN [14] is an unsupervised learning algorithm that identifies clusters in large spatial datasets by examining the local density of data points. The algorithm operates by categorizing data points into three types: core points, border points, and noise. Core points are defined by having a minimum number of neighbors within a specified epsilon (ε) radius. Border points are defined by not reaching the minimum threshold of neighbors but being within an epsilon of a core point. Finally, points that do not belong to any cluster are labeled as noise. This algorithm enables a definition of clusters with arbitrary shape and size, this unique property results in effectively grouping densely packed areas and distinguishing these from the less dense regions.
- OneClass SVMs [15] are a specialized form of SVMs designed for anomaly detection in unsupervised learning scenarios. This method works by mapping input data into a high-dimensional feature space using a kernel function, e.g. RBF or a polynomial function. Then the hyperplane that maximally separates the one class data from the origin of the transformed feature space is determined. This effectively creates a decision boundary to classify future points as part of the "normal class" or outliers. One of the practical advantages of this approach is that it doesn't require any prior knowledge of the anomalies.
- Autoencoders [16] are neural networks (NN) trained using an auto-regressive task. These NN have two main components, an encoder and a decoder. The encoder converts an input signal into a lower-dimensional latent space. The decoder then takes this compressed representation and is trained to reconstruct the original signal, hence making this an unsupervised learning method. Backpropagation is used to minimize the reconstruction error (the difference between the output and the original signal). Theoretically, a less noisy pattern takes fewer parameters to encode. As a result, the reconstruction error for these data points would be lower than those of the noisy samples, making this versatile algorithm for noise detection.

## 2.2 Data Imputation

The study subject that aims to provide solutions for missing data scenarios is the imputation field, this field started by improving on simple techniques for handling missing data (e.g., removing data points, removing features, using mean or median) which often produce biased results [17] and can now be subdivided into two large categories, statistical imputation and machine learning-based imputation.
No single imputation method consistently outperforms the rest. The effectiveness of a method depends on various factors, including the type of data, the pattern of missing values, and the specific constraints of the study [18]. This variability is underscored by numerous benchmark studies in the field, which often reach different conclusions when applied to different datasets. Jäger et al. [4] found random-forest–based imputation to be the most performant, Lin et. al. [19] concluded that deep learning methods are better across a varied selection of UCI-ML datasets with sizes ranging from 100s to 10000 data points, Sun et. al [20] found that MICE and random-forest-based methods are better for data with limited sample size (i.e., n<30k), while deep learning methods are better otherwise, and Jadhav et al. [21] found merit in using kNN for the data being tested.

Given the wide array of performance results obtained from each method depending on datasets we elected to test all state-of-the-art methods previously mentioned, including additional techniques often used in imputation. The following is a comprehensive list of the methods tested in this work for comparison's sake.

- MissForest [22] is an RF-based imputation method that has most of the properties that random-forest models have. For example, the ability to capture non-linear relationships between variables, or its tendency to perform "out-of-the-box" without requiring almost any hyperparameter optimization.
- Multivariate Imputation by Chained Equations (MICE) [23] is a technique that uses a multiple imputation approach. Much like ensemble methods in noise detection, predicting multiple imputations can reduce the risk of biased estimates and be used to calculate uncertainty.
- Multiple Imputation using Denoising Autoencoders (MIDAS) [24] [25], this is a deep learning approach where autoencoders are trained multiple times to reconstruct the dataset from different partially corrupted versions, the final imputed values will then be an average of the imputations made by each of the autoencoders.
- HyperImpute [26], is a recent state-of-the-art method that applies a generalized iterative imputation approach with an automatic model selection mechanism. This approach enables the method to adaptively select the most appropriate models for each feature based on the dataset's specific characteristics, resulting in improved accuracy.

- Deep Regressor/Classifier represents another group of deep learning-based techniques. Much like RF, it can capture non-linear patterns, but unlike random forests, deep learning techniques have been shown to utilize very large datasets more effectively. These techniques also benefit from the ability to customize the architecture and dimensionality while accounting for the properties of the dataset or the missing variable.
- Weighted K-Nearest Neighbours (kNN) [27] is a "locally grounded" method, meaning each imputation depends only on the respective neighbors, this makes it more appropriate than global imputation methods for datasets with strong local patterns. In the early 2000s kNN was often a top-performing method but in recent years as more advanced ML and other methods have evolved, kNN has been consistently outperformed. Some of the reasons for this are its inability to manage outliers, difficulty in high data dimensionality scenarios and the method tends to only be appropriate in scenarios where data is missing at random (MAR).
- SoftImpute [28] [29] is an algorithm where the user defines a soft-threshold λ and applies a dimensionality reduction technique, to the observed entries of the matrix with missing data, called singular value decomposition (SVD), values from the low-rank representation (dimensionality reduced matrix) are then used to fill in the original matrix. This process is done iteratively, with each iteration further refining the matrix approximation.
- Matrix Factorization [30] originally gained prevalence as an algorithm for recommender systems that outperformed kNN for that specific type of data. This method works by using stochastic gradient descent (SGD) to decompose a matrix with missing values into two lower-dimensional matrices. This factorization step aims to capture underlying patterns or features in the data such that when multiplied, the low-dimensionality matrixes approximate the original matrix, effectively filling in the missing values.

For kNN and MICE, we used the code implementations available in the Scikit-learn package [31]. For applying the MissForest algorithm we utilized the implementation from "missingpy" package, available on GitHub. For the deep learning regressor, we defined a simple dense network with one hidden layer and a rectified linear unit. To apply the SoftImpute and Matrix Factorization algorithms we used the implementations available at the python package "fancyimpute" [32]. For the MIDAS application we used the MIDASPy package [33].

It is important to note that every method in the current state-of-the-art requires some initial partial data from which to infer the distribution of the variable to be imputed, and usually a somewhat high percentage of that variable. In other words, none of these methods indicate a pathway to address the problem of confounding variables. Furthermore, the current methods tend to fail when a high percentage of the variable is missing, as we will demonstrate in our experiments. These are limitations that GGH addresses, using a minor setup of prior knowledge (which can be prepared in minutes) it can produce different classes of hypotheses which will be clustered based on enriched gradients ($\nabla f+$), then using as little as one data point per hypothesis class it can determine if this datapoint corroborates the overall hypothesis cluster or not. Because this methodology focuses on the distribution of enriched gradients it can be used to find correct hypotheses, but also noisy data points, which are treated as an extension of incorrect hypotheses. In other words, the method tackles both the imputation challenge as well as the noise detection challenge and can benefit from advancements in the unsupervised clustering field. Furthermore, improvements in enriched gradient representations also provide an opportunity to simultaneously improve on both tasks.

When considering scenarios where data about a specific variable is completely missing, a strategic data production effort could be used to obtain small percentages of overall complete data thus providing the requirements for the GGH framework to address confounding variables.

## 3. Methods

Data scientists with domain expertise often have knowledge beyond the features described in the data set. For example, a group of researchers might know that the scale of a chemical reaction is correlated to the experimental noise on the dependent variable [34], or that an independent variable like temperature, despite not consistently being captured in a patent/literature database, is still influencing the reaction outcome. Machine learning methods across different industries and research fields can benefit from the introduction of this knowledge but given the current architecture of deep learning models, the integration of additional unrecorded information is not trivial.

Our work aims to provide a method to easily use prior knowledge of confounding or partially missing variables to reach a model with a working understanding of these influencing factors. As a result, producing models that outperform those trained solely on recorded or imputed variables.

**Data Expansion through Hypotheses Generation:** GGH begins by transforming each incomplete data entry into a series of potential hypotheses. Each hypothesis represents a plausible completion of the missing data based on historical patterns and available complete rows.

**Gradient Analysis for Hypothesis Evaluation:** Once the hypotheses are generated, GGH utilizes a novel gradient-based technique to evaluate them. During the training of the ML model, gradients are computed not only for the existing complete data but also for each generated hypothesis. These gradients are enriched with additional information and will be referenced as $\nabla f+$. Since the ground truth should follow a subset of consistent patterns while the incorrect hypotheses should follow subsets of diverse patterns, the latent representation of the enriched gradients should be distributed differently across the latent space. With $\nabla f+$ for the correct hypothesis being more constricted than the incorrect ones. It is then possible to use a density-based clustering algorithm to distinguish noisy data points or train a OneClass SVM on the $\nabla f+$ of the ground truth data to identify correct hypotheses. A selection radius parameter that selects gradients around ground truth for backpropagation should be optimized depending on the amount of available ground truth data.

### 3.1 Gradient Guided Hypotheses

In Figure 1, there is an illustrated example of a dataset with four independent variables where $V_4$ contains a significant portion of its values missing. The scientist applying GGH should define a set of class values that $V_4$ can reasonably assume in a general distribution. This set of values is then used to automatically expand instances with missing $V_4$ values into all possible complete class hypotheses that the missing variable could have assumed. Even though a class hypothesis might not have the exact value the missing variable had, the algorithm will work if a sufficiently close class can be established. Simultaneously, if there are data points containing complete rows, these are separated and both the hypothesis group as well as the complete rows group go through a forward pass on an ML model. Because the algorithm requires the use of enriched gradients the ML method should utilize SGD for convergence. Aside from this requirement, this technique is very versatile because it is architecture and task-agnostic. Furthermore, improvements to the machine learning methods being used should benefit the quality of the hypothesis clusters, as more advanced models will be able to detect higher complexity patterns influencing the distribution of the gradients in the latent space.

After execution of the forward pass, the individual loss ($H_L$, $G_L$) for each instance of the hypotheses (H) and ground-truth (G) are calculated, as well as the respective gradients ($H_{\nabla f}$, $G_{\nabla f}$).

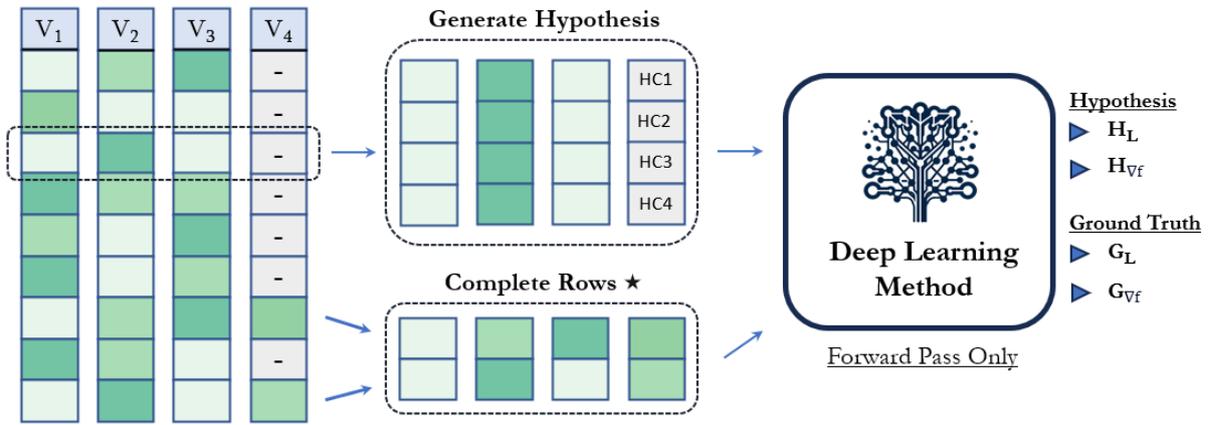

Figure 1 – Schematic representation of dataset being separated into complete and incomplete rows. With incomplete rows being expanded into multiple hypothesis, Hypothesis Class 1 (HC1), HC2, HC3 and so forth. Both batches of data then go through a forward pass of a method utilizing SGD for learning, and individual loss ($H_L$, $G_L$) and gradients ($H_{\nabla f}$, $G_{\nabla f}$) are calculated for each hypothesis ($H_i$) and each complete row ($G_i$).

During an initial warmup phase defined by the number of epochs until a η parameter, all calculated gradients are selected for backpropagation. Due to the total number of hypotheses gradients, these can often overwhelm the signal in $G_{\nabla f}$, for this reason, $G_{\nabla f}$ are amplified by an α parameter. After the warmup phase the $\nabla f$, can optionally undergo a signal processing stage. The $\nabla f$ depend in part on the input data, and depending on feature importance most of this signal will come from the fully captured variables (complete columns), meaning that $\nabla f$ from the same hypothesis class will be distributed differently in the latent space depending on the non-hypothesized input signal. Generally, this should generate consistent patterns, which would then differ depending on the hypotheses, but in some scenarios, it might be useful to attempt to decouple the gradient signal associated with the hypothesis from the gradient signal associated with the rest of the instance data. To achieve this the algorithm selects the gradients from hypotheses generated for the same instance and removes the average gradients from the group, since the gradient signal from the non-hypothesized input is the only constant across the group, this will be strongly represented in the average of the group. This approach does introduce some signal contamination from other hypotheses.

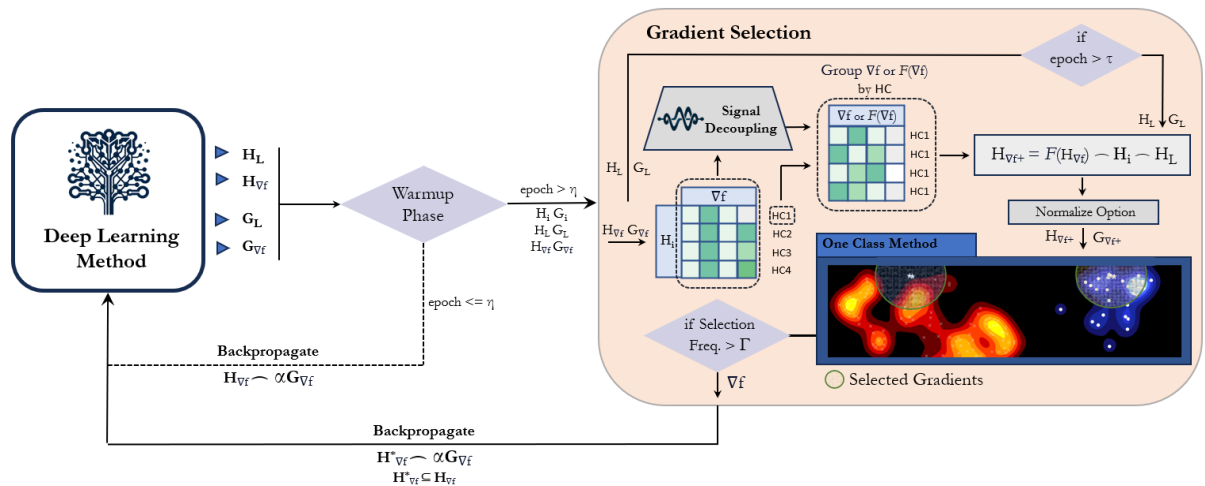

Figure 2 - Schematic representation of the GGH algorithm. After expanding the dataset with hypotheses and executing a forward pass as shown in Figure 1. During the warmup phase, all hypothesis and ground truth gradients are used for backpropagation. After the warmup phase gradients ($\nabla f$) from the hypotheses are grouped by source data, and can optionally undergo a signal decoupling step. The resulting gradients are regrouped by the hypothesis class and enriched with source data and loss values after the staggering

period is concluded. The resulting groups of vectors can then optionally be normalized before using enriched gradients from ground truth ($G_{\nabla f+}$) to train the One Class Method, which will then select $H_{\nabla f+}$ according to proximity to $G_{\nabla f+}$ in the latent space. Finally, a high-pass filter is applied such that only gradients consistently selected across all epochs up that epoch are selected for backpropagation through the deep learning model.

After grouping $\nabla f$ by instances and processing the $\nabla f$, these will be grouped by hypothesis class, to save on compute, only gradients for the second to last layer are used, and these are enriched with instance input data, as well as the previously calculated $H_L$, $G_L$. Loss is an interesting and perilous feature, its introduction into the $\nabla f+$ vector is inspired by the use of the reconstruction error in Autoencoders when detecting noisy data. If noise is a collection of scattered patterns while truth underlines a consistent set of patterns, then the loss for correct hypotheses should be lower, than for incorrect ones, however, loss also depends on model weights, so backpropagating on hypotheses with lower loss can cause a self-reinforced feedback loop, where weights are updated to reduce error for those hypotheses further decreasing the loss and hence increasing the chances of these hypotheses being selected again. To counter this effect, the introduction of the $H_L$, $G_L$ into the $H_{\nabla f+}$ and $G_{\nabla f+}$ is staggered, being introduced after $\tau$ epochs. Finally, the Loss generated from each hypothesis also acts as an indirect proxy for how often the $\nabla f$ of the hypothesis have been selected and backpropagated on.

The $\nabla f+$ can then be normalized per hypothesis class batch and the $G_{\nabla f+}$ are used to fit a one-class model, in this case, we used OneClass SVM, this model then estimates which instances of $H_{\nabla f+}$ follow the same patterns as those fitted from $G_{\nabla f+}$. If the instances selected from $H_{\nabla f+}$ are not present in the exclusion list created from the hypothesis of G and have been selected ratio of time higher than Γ, a parameter which defines the minimum frequency ratio, the $\nabla f$ from these hypotheses will be sent to the machine learning model to be backpropagated upon.

## 3.2 Enriched Gradients Distribution

To demonstrate how the algorithm is functioning we use a real-world open-source dataset from a study into the stability of organic photovoltaics [35]. This dataset reports the degradation of polymer blends for organic solar cells under exposure to light. To test the method as a solution for missing data, we separate the data into train-validation-test, and hide 98.5% of the column containing the ratio of a polymer layer PCBM. After 40 epochs of training a DL model with GGH, we select the enriched gradients for the last batch of hypotheses where PCBM hypothesized ratio in the blend is 60%. Under normal circumstances, all $H_{\nabla f+}$ are estimated in an unsupervised scenario, meaning it wouldn't be known a priori which hypotheses are correct, but for the sake of demonstration, we separate $H_{\nabla f+}$ into correct and possibly incorrect hypotheses, using the complete dataset, as reference. In Figure 3, a tSNE plot shows how $H_{\nabla f+}$ generated from possibly incorrect hypotheses are distributed in the space, as well as the $G_{\nabla f+}$ the model had access to, Figure 4 shows the same plot but for the $H_{\nabla f+}$ generated from the correct hypothesis.

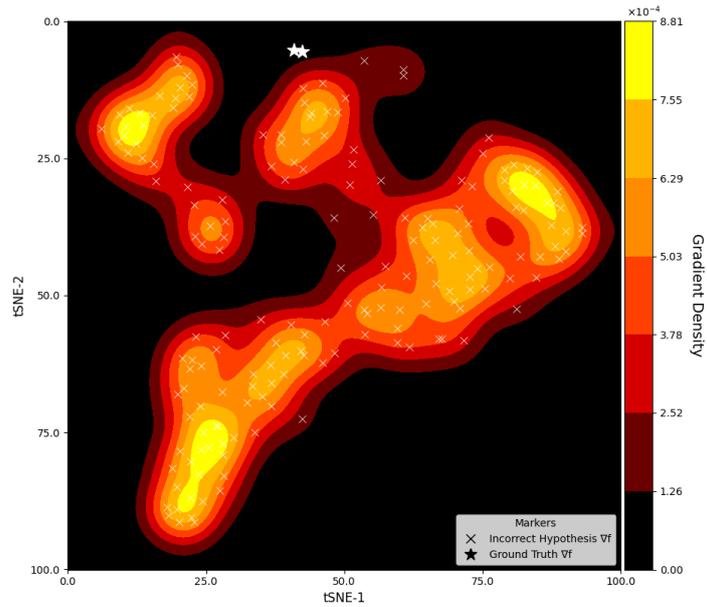

Figure 3 – tSNE visualization by dimensionality reduction of enriched gradients vectors of potentially incorrect hypotheses and ground truth instances. Ground truth is defined as instances where no variable had missing data. Colour represents point density in the space.

When comparing Figure 3 to Figure 4, it's clear that the $H_{\nabla f+}$ from potentially incorrect hypotheses generate a lot more clusters that span across the latent space, while $H_{\nabla f+}$ from correct hypotheses cluster and concentrate in a specific section of the latent space. This is the property that GGH aims to enhance and capitalize on, to successfully distinguish correct hypotheses from incorrect ones. From Figure 4 it is also elementary to comprehend why even just a single example of the ground truth data for a specific hypothesis class, in this scenario, can suffice to select a high ratio of correct hypotheses to backpropagate on, a characteristic that is not present in any other state-of-the-art imputation method.

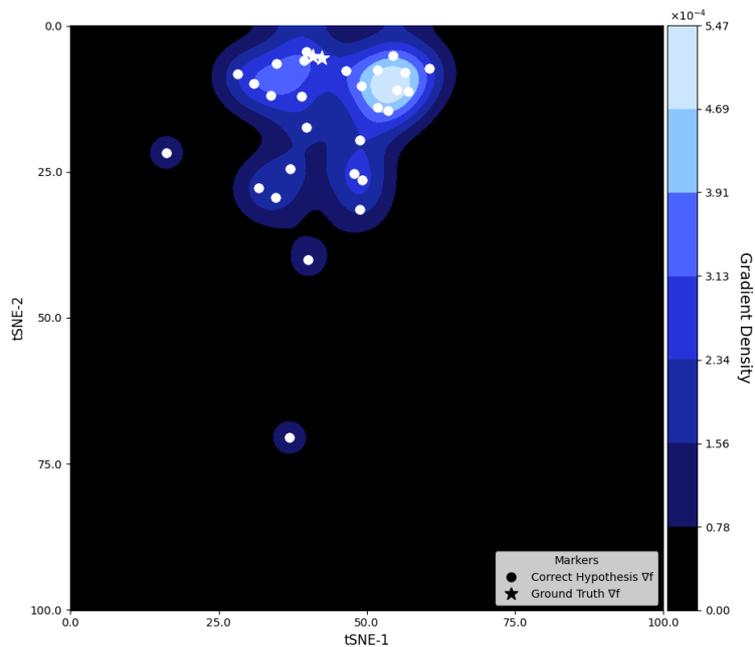

Figure 4 - tSNE visualization by dimensionality reduction of enriched gradients vectors of correct hypotheses and ground truth instances. Ground truth is defined as instances where no variable had missing data. Colour represents point density in the space.

A clear way to mentally extrapolate the operations that the OneClass SVM model is executing is to imagine the RBF kernel function defining a region in the high-dimensionality space of the ∇f+ which essentially corresponds to a circle drawn around the ground truth represented as stars in these visualizations. Such that other ∇f+ that fall inside this circle will be selected for backpropagation. The "radius of this circle" is an important parameter in GGH which should be defined according to the amount of ground truth data and complexity of the underlying patterns. In the OneClass SVM, this "radius" can be defined through the hyperparameter nu.

We denominate the instances that generated the ∇f+ in Figure 3 as a potentially incorrect hypothesis. This is because in scenarios where multiple values of the missing variable would lead to the same outcome, we can have multiple hypotheses generated for the same instance which are correct, but when evaluating the selection only one hypothesis is covered in the data. This means when the GGH selects potentially incorrect hypotheses what the method is doing is selecting ∇f+ which are similar to $G_{\nabla f+}$. Because these are similar it does not cause the model to diverge as much as if a distant incorrect $H_{\nabla f+}$ was selected instead. In other words, there are different types of potentially incorrect hypotheses, some are correct, some are irrelevant, and some cause the model to completely diverge from the ground truth. If the full incorrect hypotheses distribution of the gradients matched that of the correct complete gradients, then the resulting model trained solely on the incorrect hypothesis would generally converge to the same endpoint as a model trained in ground truth.

### 3.3 Hypotheses Selection Probability

The distribution of ∇f+ will depend on the data, model architecture, parameter initialization, hypothesis class and will change with each weight update. So, while Figure 3 and Figure 4 are useful to analyze how the ∇f+ are behaving depending on different transformation techniques that generate ∇f+, an illustration that combines the performance of the transformation techniques with the quality of the selection technique is necessary. Figure 5 aims to provide such a visual, as a proxy to how successful ∇f+ generation and selection were simultaneously, by presenting the probability distribution of correct and potentially incorrect hypotheses to be selected during the training of a model. In a scenario where ∇f+ from correct hypotheses would cluster perfectly around G$\nabla_{f+}$ and where the selection algorithm correctly selects the entire cluster the green distribution should be completely skewed to the right, with most or all correct hypotheses being selected consistently in epochs after the warmup phase. In such a scenario, we would expect the potentially incorrect hypothesis to present the shape of a U skewed towards the left, with some of those which are actually correct on the far right.

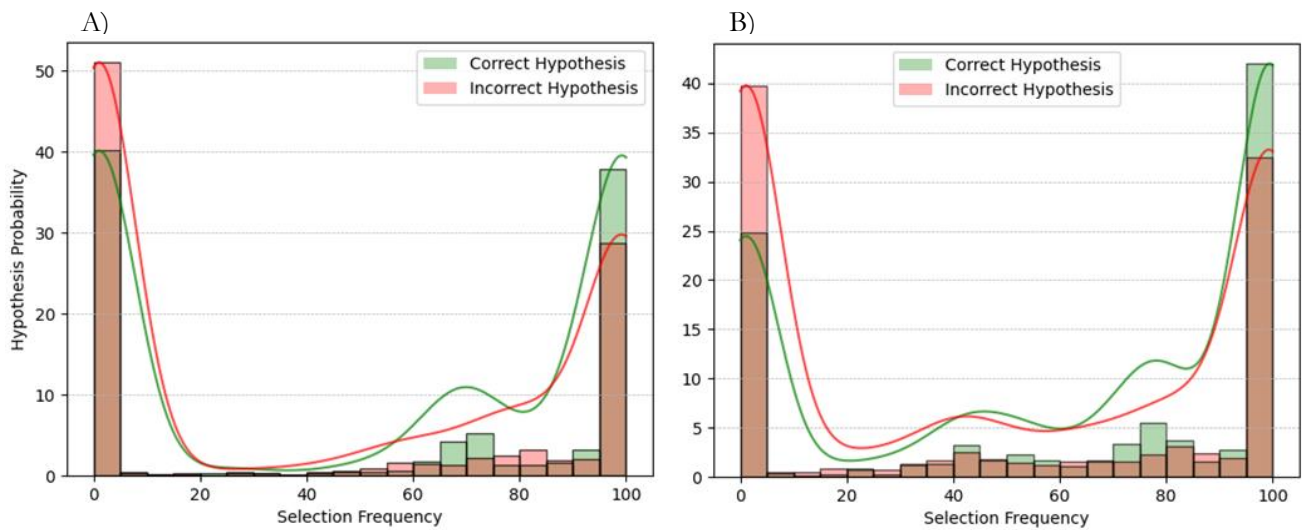

Figure 5 – Histograms showcasing the probability distribution of the ∇f+ of correct and potentially incorrect hypothesis after 100 epochs. A) presents these distributions when 98.5% of an independent variable was missing, while B) presents the distributions for when 97% of the same variable was missing. With more data on the incomplete variable (B) plot), the gap between the probability of selecting a potentially incorrect and correct hypothesis for training the model increases.

Figure 5 also showcases what happens as the method has access to more ground truth data points of the incomplete variable, plot A) showcases the distributions when 98.5% of the variable was missing and B) 97%. In this case, the nu parameter (selection region "radius") of OneClass SVM was constant, meaning that the increase in ground truth led to an increase of regions of the latent $\nabla f+$ space being selected, this then means that the probability of selecting hypotheses increases which is represented as a shift of both distributions towards to right. Since the distribution of G$\nabla_{f+}$ in the latent space is closer to $\nabla f+$ from the correct hypothesis, this probability distribution shifts more than the distribution for the potentially incorrect hypotheses.

### 3.4 Noise Detection

One of the advantages of GGH formulation is that it allows to apply the same principles and techniques to both data imputation and noise detection challenges. Depending on the model and amount of noise, the weights of a model training on a noisy dataset will converge to a different endpoint than the weight of the same model training on clean data. Because the convergence endpoint is different, there's a high probability the gradients generated for each group are different as well, so the $\nabla f+$ will cluster differently. This knowledge can be utilized to specifically select the gradients of less noisy data points for backpropagation.

The following visualizations presented in Figure 6 and Figure 7 were generated using the Photocell Degradation [35] dataset. In Figure 6 it is possible to observe how $\nabla f+$ generated from noisy data points cluster on the edges of the space, while $\nabla f+$ from unaltered data points cluster in the center of the space. This topology allows the use of unsupervised density clustering methods like DBSCAN to be used to separate the two groups.

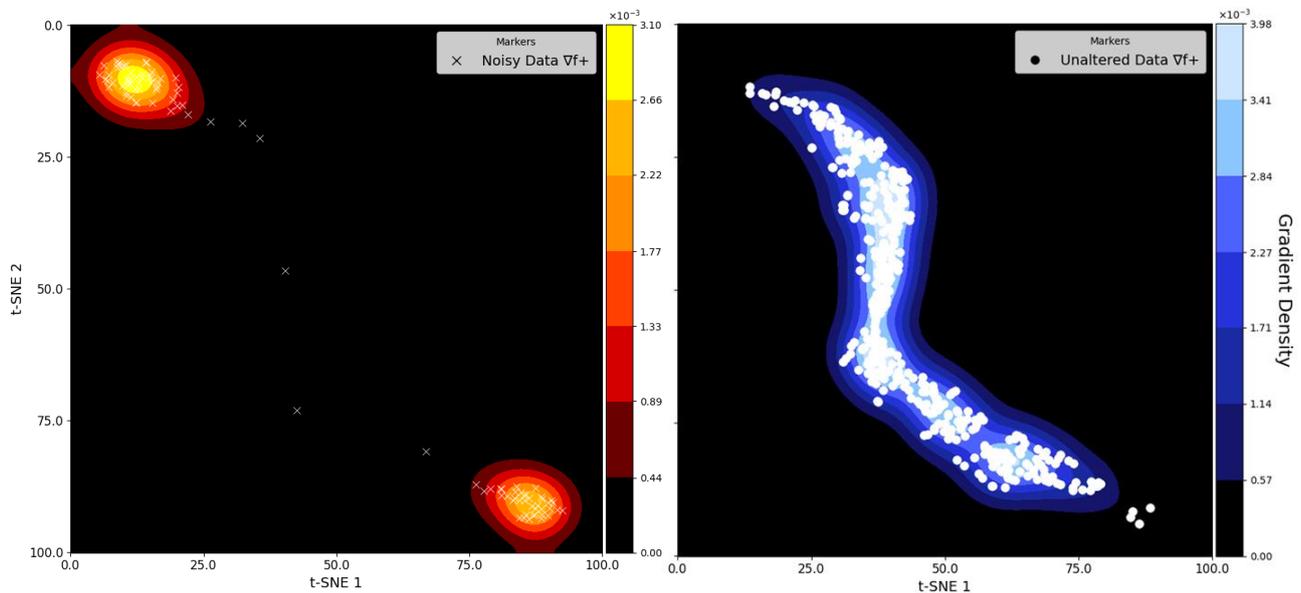

Figure 6 - tSNE visualization by dimensionality reduction of enriched gradients vectors with colour representing point density in the space. A) Contains $\nabla f+$ calculated after 200 epochs from data points on which 20 to 40% of the target variable range is added as noise. B) Contains $\nabla f+$ calculated after 200 epochs from data points which are unaltered.

In Figure 7, we present the confusion matrix resulting from using GGH after 200 epochs of selecting gradients from non-noisy data points to train, from this confusion matrix it's possible to observe that when the method estimates a datapoint to have noise, its precision is 93.75% with overall accuracy at 94.4%.

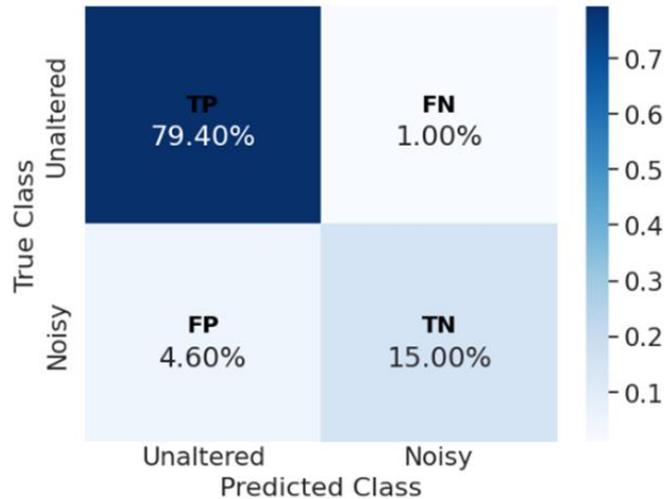

Figure 7 - Normalized confusion matrix, for unsupervised noise labelling using density-based clustering method.

## 4. Results

The visualizations previously shown in the manuscript give insights into how the different components of the algorithm are functioning. In this section, we validate GGH by comparing how it performs against current state-of-the-art imputation methods in different real-world datasets and scenarios. We also compare the performance of a model when trained with GGH as an "on the fly" filter for noisy data versus training on the data without applying noise detection.

GGH is compared on the test set against two baseline techniques and the best-performing imputation method for all datasets. The baseline techniques are "Complete Columns" and "Complete Rows". When a very high percentage of a variable is missing, (eg, 98.5%) it is common in the field of data science to drop this column, since it is expected that imputation will not be a worthy endeavor, since in the open-source datasets we used, the rest of the columns are complete, this technique is denominated in the results tables as "Complete Columns". In scenarios where the variables with missing data are important and should not be dropped, another common technique is to keep only the rows that have no missing data, denominated in the results tables as "Complete Rows".

For the imputation, we tested the eight methods that are discussed in the 2.2 Data Imputation section of the manuscript. The "Best Imputation Method" presented in the results tables, is the average across test splits of the best performing imputation method across the average of the validation splits.

For all methods on the results tables early stopping is used to select the best checkpoint from each particular method for each validation split, this checkpoint is then used to estimate the test.

### 4.1 Photocell Degradation

To validate the efficacy of GGH as a method to address variables with missing data, we selected an open-source dataset [35], holding records of the degradation of polymer blends for organic solar cells under exposure to light. To test the method as a solution for missing data, we separate the data into train-validation-test and hide 98.5% of the column with the ratio of a polymer layer PCBM, the results for the comparison are presented in Table 1. The initial dataset contained approximately 750 datapoints in the trainset, the PCBM ratios fall within one of 6 hypothesis classes, giving on average 125 values per hypothesis class. This means that when simulating that 98.5% of the values are missing there are on average only 1.87 ground truth data points per hypothesis class in the training set.

When simulating a scenario with such a high ratio of missing data, we observe that applying current state-of-the-art imputation methods is comparable to using the baseline techniques, this is because these methods require considerable sampling of the distribution of the missing variable to correctly fill in the missing values. Because of the underlying algorithm of GGH, it's possible that as little as one data point per Hypothesis Class can already be used to identify the region of the pattern space that should be used to backpropagate. However, for the current version of GGH there is still a reliance on ground truth examples, and when sampling a few data points randomly from its distribution, these ground truth data points can be representative or not of the overall distribution, hence the high standard deviation associated with applying GGH in such high ratios of missing data as observed in Table 1.

| **Method** | **R2 Score** | **MSE**$(\times 10^{-2})$ | **MAE**$(\times 10^{-2})$ |
|---|---|---|---|
| Complete Columns | 14.3 $\pm$ 4.1 | 1.1 $\pm$ 0.2 | 7.7 $\pm$ 0.5 |
| Complete Rows | 15.9 $\pm$ 17.6 | 1.1 $\pm$ 0.3 | 7.8 $\pm$ 1.0 |
| Best Imputation Method | 17.9 $\pm$ 11.2 | 1.1 $\pm$ 0.2 | 7.5 $\pm$ 0.8 |
| **GGH** | **32.5 $\pm$ 19.8** | **0.9 $\pm$ 0.3** | **6.8 $\pm$ 1.1** |
| Complete Data | 88.0 $\pm$ 3.2 | 0.2 $\pm$ 0.0 | 2.6 $\pm$ 0.3 |

Table 1 - Average performance of each method on the test set across 15 separate randomized runs for the Photocell Degradation[35] regression task, with 98.5% of an important independent variable missing. For all methods presented in the table, a first train and validation step are executed, with the best model on the validation set being the one selected to estimate the test. Complete columns and complete rows as the names imply refer to training the neural network with subsets of the dataset, where either all columns or all rows have no missing data. The "Best Imputation Method" represents the model performance after using the best SOTA method according to performance on the validation set to impute the data.

In this scenario with very high ratios of missing data GGH represents the only viable solution, showing a large improvement over the state of the art. Such scenarios are not implausible, in fact, these could be manufactured as a novel approach to tackle unrecorded variables, because the data requirements are so low, small data production efforts could be executed to attain a few records on the unrecorded or highly missing variable.

The principles of gradient enrichment and gradient selection can also be applied to detect very noisy data points and to avoid backpropagating the gradients derived from these. The unsupervised method does differ from the one used for missing data, which used OneClass SVM, and GGH selects gradients based on a density-based clustering algorithm instead. To test the viability of using enriched gradients for distinguishing noisy data from normal data, a simulation of noise with the level of 40% to 60% was applied in 30% of the train data, GGH was then used to filter out the noisy data points and a model was trained on the resulting dataset. The results comparing the performance of the model on the clean data, noisy data and filtered noisy data can be observed in Table *2*.

| Method | R2 Score | MSE (x10⁻³) | MAE (x10⁻²) |
|---|---|---|---|
| Complete+Noise | 67.4 ± 8.0 | 3.9 ± 1.0 | 4.5 ± 5.7 |
| **GGH Noise Filter** | **79.9 ± 6.9** | **2.4 ± 0.9** | **3.1 ± 0.4** |
| Clean Data | 84.5 ± 5.1 | 1.8 ± 0.5 | 2.7 ± 0.3 |

Table 2 - Average performance of each method on the test set across 15 separate randomized runs for the Photocell Degradation regression task, with simulated noise of 40% to 60% deviation on the dependant variable in 30% of the train data. For all methods presented in the table, a first train and validation step are executed, with the best model on the validation set being the one selected to estimate the test.

It is important to note that depending on the data, the enriched gradients from outliers will cluster differently from the majority of the train dataset, and can easily be mistaken by noise in such an approach, to address this issue, once the model converges with the filtered data, we run additional epochs with all the train data. By doing this, we observe a contraction of the space where all clusters become closer, importantly the clusters of $\nabla f+$ of the outliers approach the cluster of the remaining clean data in fewer epochs, thus separating from the $\nabla f+$ of the noisy data points.

## 4.2 Photoredox Yield Estimation

For the second validation use case, we used a published high-throughput experimentation dataset of photoredox C–N coupling chemical reactions [36]. This is an interesting high-quality dataset to test GGH in a synthesis modeling use case, because in addition to containing molecular structures, it also captured molecular ratios with high granularity for each reaction. Even when training large language models on millions of chemical reactions represented as SMILES, the performance of these models is limited by available complete data, with condition and procedure data missing from the modeling having a large impact on train-test with high-quality datasets [37]. With GGH, it is possible to generate hypotheses for the missing condition, eg molecular ratios, and learn about the impact of molecular ratios in highly incomplete synthesis data.

To simulate such a scenario, we mask 70% of the molecular ratios for the photocatalyst and train a model a regression model to estimate LC-MS, a proxy for product yield. Reactivity in photoredox C–N coupling reactions involves complex molecular interactions, this complexity is magnified by varying molecular ratios. In this use case, even when utilizing 30% of the photocatalysts molecular ratios, the best imputation method and baseline techniques perform similarly, once again GGH performs significantly better, enabling the model to learn the impact of this highly complex and incomplete variable, as observed in Table 3.

| Method | R2 Score | MSE (x10⁻²) | MAE (x10⁻²) |
|---|---|---|---|
| Complete Columns | 68.4 ± 7.3 | 1.6 ± 0.4 | 7.0 ± 1.0 |
| Complete Rows | 69.7 ± 5.6 | 1.5 ± 0.3 | 7.2 ± 0.9 |
| Best Imputation Method | 68.5 ± 9.5 | 1.6 ± 0.6 | 7.6 ± 1.6 |
| **GGH** | **75.0 ± 5.3** | **1.3 ± 0.3** | **6.8 ± 0.9** |
| Complete Data | 81.4 ± 4.5 | 0.9 ± 0.3 | 5.4 ± 0.7 |

Table 3 - Average performance of each method on the test set across 15 separate randomized runs for the Photo-Redox Yield Estimation task, with 70% of the values for photocatalyst equivalent missing. For all methods presented in the table a first train and validation step are executed with the best model on the validation set being the one selected to estimate the test. Complete columns and complete rows as the names imply refer to training the neural network with subsets of the dataset, where either all columns or all rows have no missing data. The "Best Imputation Method" represents the model performance after using the best SOTA method according to performance on the validation set to impute the data. GGH is shown to perform significantly better than all other methods.

As in the previous noise benchmark, the same simulation of noise with the level of 40% to 60% was applied in 30% of the train data, once again GGH was able to detect noisy data points leading to improved performance once these were filtered out, as shown in Table 4. In this dataset, it was especially important to have a distinction between outliers and noisy data points, as the clusters for these ∇f+ overlapped after the initial filtering and convergence. As already explained, this was achieved by a second training stage where no selection is done and these clusters present different approximation rates to ∇f+ of clean data points.

| Method | R2 Score | MSE (x10⁻³) | MAE (x10⁻²) |
| --- | --- | --- | --- |
| Complete+Noise | 60.8 ± 8.6 | 19.3 ± 3.7 | 10.0 ± 0.8 |
| **GGH Noise Filter** | **67.4 ± 5.7** | **16.3 ± 3.6** | **8.5 ± 1.0** |
| Complete Data | 85.8 ± 4.0 | 7.1 ± 2.0 | 4.1 ± 0.5 |

Table 4 - Average performance of each method on the test set across 15 separate randomized runs for the Photo-Redox Yield Estimation task, with simulated noise of 40% to 60% deviation on the dependant variable in 30% of the train data. For all methods presented in the table a first train and validation step are executed with the best model on the validation set being the one selected to estimate the test.

### 4.3 Airfoil Self-Noise

The third dataset used to benchmark the method against other state-of-the-art approaches is called Airfoil Self-Noise. This data set was made available by NASA at the UCI ML repository. It was obtained from a series of aerodynamic and acoustic tests of two and three-dimensional airfoil blade sections conducted in an anechoic wind tunnel. The goal is to perform a regression task to estimate the scaled sound pressure level in decibels.

For this dataset 96% of an important variable, chord length, was simulated to be missing at random. Once again, the current state-of-the-art imputation methods that were tested did not perform better than the baselines, indicating its inability to function with a high percentage of MAR data.

| Method | R2 Score | MSE (x10⁻²) | MAE |
| --- | --- | --- | --- |
| Complete Columns | 52.9 ± 9.1 | 1.6 ± 0.3 | 10.0 ± 1.0 |
| Complete Rows | 39.9 ± 17.0 | 2.0 ± 0.4 | 10.9 ± 1.2 |
| Best Imputation Method | 52.7 ± 10.0 | 1.6 ± 0.3 | 9.7 ± 0.9 |
| **GGH** | **68.1 ± 9.5** | **1.1 ± 0.3** | **8.0 ± 1.2** |
| Complete Data | 79.6 ± 5.4 | 0.7 ± 0.2 | 6.0 ± 0.8 |

## 5. Discussion

The Gradient Guided Hypotheses (GGH) method addresses critical challenges in handling missing and noisy data, applicable across various scientific and industrial domains. By generating hypotheses for missing data and distinguishing noisy data through gradient analysis, GGH enhances both data quality and model performance. Particularly in extremely scarce data regimes such as 70-98%, GGH outperformed all state-of-the-art methods. Its architecture-agnostic nature allows for broad application without needing modifications to existing machine learning frameworks, and its ability to operate effectively with minimal ground truth data can significantly reduce data collection costs and efforts. Implementing automatic hyperparameter tuning techniques such as Bayesian optimization could substantially simplify the deployment of GGH, making it more accessible and effective, for different scarcity and data type scenarios.

Furthermore, current applications suggested in this study focused on one variable missing regime, which presents an opportunity for future work to address a scope increase of the number of missing variables. The algorithm was demonstrated using a dense network as the machine learning model from where the gradients are calculated, we anticipate that with the code optimized to utilize specialized more advanced models, the gradients would capture more informative patterns further boosting the overall performance of the method.

## Availability of data and materials

Data and code are available at https://github.com/schwallergroup/gradient_guided_hypotheses

## Acknowledgments

PS acknowledges support from the NCCR Catalysis (grant number 180544), a National Centre of Competence in Research funded by the Swiss National Science Foundation.